\documentclass{article}

\usepackage{microtype}
\usepackage{graphicx}
\usepackage{subcaption}
\usepackage{booktabs} % for professional tables
\usepackage{multirow}
\usepackage{hyperref}

\usepackage[preprint]{icml2026}

\usepackage{amsmath}
\usepackage{amssymb}
\usepackage{mathtools}
\usepackage{amsthm}
\usepackage{subcaption}
\usepackage[percent]{overpic}
\usepackage{listings}
\usepackage[capitalize,noabbrev]{cleveref}

\theoremstyle{plain}

\theoremstyle{definition}

\theoremstyle{remark}

\usepackage[textsize=tiny]{todonotes}

\icmltitlerunning{BatCoder: Self-Supervised Bidirectional Code-Documentation Learning via Back-Translation}

\begin{document}

\twocolumn[
  \icmltitle{BatCoder: Self-Supervised Bidirectional Code-Documentation 
  \\
  Learning via Back-Translation }



  % 这里的 equal 对应文末的 Equal contribution 脚注
  \icmlsetsymbol{equal}{*}

  \begin{icmlauthorlist}
    % 所有人都来自 fudan
    \icmlauthor{Jingwen Xu}{equal,fudan}
    \icmlauthor{Yiyang Lu}{equal,fudan}
    \icmlauthor{Zisu Huang}{fudan}
    \icmlauthor{Changze Lv}{fudan}
    \icmlauthor{Xiaohua Wang}{fudan}
    \icmlauthor{Shizheng Li}{fudan}
    \icmlauthor{Zhibo Xu}{fudan}
    \icmlauthor{Zhengkang Guo}{fudan}
    \icmlauthor{Zhengyuan Wang}{fudan}
    \icmlauthor{Muzhao Tian}{fudan}
    \icmlauthor{Xuanjing Huang}{fudan}
    \icmlauthor{Xiaoqing Zheng}{fudan}
  \end{icmlauthorlist}

  % 定义复旦大学的单位信息
  \icmlaffiliation{fudan}{College of Computer Science and Artificial Intelligence, Fudan University, Shanghai, China}

  % 共一的邮箱与通讯
  \icmlcorrespondingauthor{Jingwen Xu}{xujw24@m.fudan.edu.cn}
  \icmlcorrespondingauthor{Yiyang Lu}{yylu24@m.fudan.edu.cn}
  \icmlcorrespondingauthor{Xuanjing Huang}{xjhuang@fudan.edu.cn}
  \icmlcorrespondingauthor{Xiaoqing Zheng}{xqzheng@fudan.edu.cn}

  \icmlkeywords{code, back-translation, self-supervised}

  \vskip 0.3in
]



\printAffiliationsAndNotice{}  % no special notice (required even if empty)

\begin{abstract}
Training LLMs for code-related tasks typically depends on high-quality code-documentation pairs, which are costly to curate and often scarce for niche programming languages. 
We introduce BatCoder, a self-supervised reinforcement learning framework designed to jointly optimize code generation and documentation production. 
BatCoder employs a back-translation strategy: a documentation is first generated from code, and then the generated documentation is used to reconstruct the original code. 
The semantic similarity between the original and reconstructed code serves as an implicit reward, enabling reinforcement learning to improve the model’s performance both in generating code from documentation and vice versa. 
This approach allows models to be trained using only code, substantially increasing the available training examples.
Evaluated on HumanEval and MBPP with a $7$B model, BatCoder achieved $83.5\%$ and $81.0\%$ pass@$1$, outperforming strong open-source baselines. 
Moreover, the framework demonstrates consistent scaling with respect to both training corpus size and model capacity.

% Training large language models for code-related tasks often relies on high-quality code-documentation pairs, which are expensive to curate and particularly scarce for many programming languages. 
% Existing approaches mitigate this limitation through data augmentation or synthetic supervision, but typically rely on stronger external models or fixed heuristics to generate training data, making the quality of synthesized data difficult to guarantee and inherently constraining further model improvement by the capacity of the generating model.
% We propose \textbf{BatCoder}, a self-supervised reinforcement learning framework for jointly optimizing code generation and documentation synthesis. The framework utilizes a back-translation process where documentation is generated from code and subsequently used to reconstruct the original program. The semantic consistency between the original and reconstructed code serves as an implicit reward, guiding the optimization of model parameters via reinforcement learning without the need for paired data or teacher models.
% On HumanEval and MBPP, BatCoder achieves 83.5 / 81.0 pass@1 with a 7B model, outperforming strong open-source baselines at comparable scales. Moreover, we observe that the improvements are more substantial for programming languages with limited training data like Ruby and Go, and that the proposed framework exhibits consistent scaling behavior with respect to both training corpus size and model capacity, indicating strong robustness and promising scalability.

\end{abstract}

\section{Introduction}
\label{sec:intro}
\vspace{-3mm}

The interaction of natural language and programming languages has been a core focus in software engineering research. Key directions include generating executable code from natural language specifications \cite{haiduc2008use, gulwani2011automating, chen2021evaluating} and producing natural language documentation from source code \cite{mcburney2015automatic, jiang2017automatically, iyer2016summarizing}. These bidirectional transformations bridge human intent and machine-executable logic, enabling rapid prototyping and code synthesis \cite{austin2021program, hendrycks2021measuring, zhuo2024bigcodebench}, while facilitating automated commenting and legacy code maintenance \cite{stolee2014solving, wan2018improving}. Overall, they enhance developer productivity and software maintainability across the development lifecycle.

Code-related tasks have been studied for over a decade.
% including generating code from natural language intent specifications \cite{haiduc2008use}, and producing natural language summaries from source code \cite{stolee2014solving,mcburney2015automatic,jiang2017automatically}. 
Early approaches relied on information retrieval techniques \cite{ye2016word} as well as statistical models \cite{iyer2016summarizing,wan2018improving}. Recent advances in LLMs have substantially improved bidirectional generation between natural language and programming code. Foundation models such as GPT-5 \cite{openai2025gpt5_card}, Qwen3 \cite{qwen3technicalreport}, Gemini 2.5 \cite{gemini2.5}, and Claude 3.7 \cite{claude3.7} demonstrate strong code understanding and generation capabilities. In parallel, code-specialized models, including CodeLLaMA \cite{roziere2023code}, DeepSeek-Coder V2 \cite{zhu2024deepseek}, and Qwen3-Coder \cite{qwen3technicalreport}, have emerged and achieved strong performance across a wide range of code-related benchmarks \cite{elnaggar2021codetrans,chen2021evaluating,austin2021program, zhuo2024bigcodebench}.
% Translating between natural language and programming languages has been actively studied in computer science. Early work explored automating the generation of code from high-level intent specifications, such as natural language descriptions \cite{haiduc2008use} or input-output examples~\cite{gulwani2011automating}, and the production of natural language summaries from source code \cite{stolee2014solving,mcburney2015automatic,jiang2017automatically}, with approaches including information retrieval \cite{ye2016word} and statistical or neural modeling \cite{iyer2016summarizing,wan2018improving}. 
% Recent progress in large language models (LLMs) has substantially improved the ability to process both natural languages and programming languages. 
% General models such as GPT-5 \cite{openai2025gpt5_card}, Qwen3 \cite{qwen3technicalreport}, Gemini 2.5 \cite{gemini2.5}, and Claude~3.7 \cite{claude3.7} demonstrate strong cross-domain reasoning and code understanding capabilities. 
% Building on these foundations, code-oriented models including CodeLlama \cite{roziere2023code}, DeepSeek-Coder~V2 \cite{zhu2024deepseek}, and Qwen3-Coder \cite{qwen3technicalreport} further specialize through domain-adaptive pretraining and instruction tuning, achieving state-of-the-art results on code translation \cite{elnaggar2021codetrans}, completion \cite{chen2021evaluating,austin2021program}, and synthesis benchmarks \cite{hendrycks2021measuring,zhuo2024bigcodebench}.

However, training LLMs for code-documentation alignment and transformation tasks, such as code description and documentation generation, typically relies on large collections of high-quality code-documentation pairs. Despite the abundance of raw source code in public repositories such as GitHub, such paired supervision remains limited and uneven in quality, which in turn constrains the scalability and generalization of models on these tasks. To address this bottleneck, prior work has explored various data augmentation strategies. 
For example, WizardCoder~\cite{luo2023wizardcoder} iteratively evolves instructions to construct diverse training pairs, while Magicoder \cite{wei2024magicoder} synthesizes coding problems from unlabeled code snippets. \citet{gao2024learning} employs self-supervised pseudo-labeling on unlabeled code. Although effective, these approaches typically require stronger external models to synthesize documentation or pseudo-labels, preventing the target model from leveraging the same mechanism for self-improvement. 
Moreover, once such code-documentation pairs are constructed, model optimization is usually carried out using conventional supervised fine-tuning (SFT) or reinforcement learning (RL) paradigms, where the generated documentation is treated as fixed supervision rather than being explicitly evaluated or optimized with respect to the training objective.

To overcome these constraints, we revisit the problem from a self-supervised perspective and explore whether meaningful training signals can be derived directly from unlabeled code via model generation and feedback. A key observation is that well-formed documentation should preserve sufficient information to enable a faithful reconstruction of the code, and effective code generation should adhere to the requirements outlined in the documentation. Consequently, the reconstructed code is expected to be similar with the original code. 
This relationship provides a natural foundation for learning code-documentation transformations without relying on explicit paired data.
This observation naturally motivates a back-translation learning paradigm, in which documentation is generated from code and then used to regenerate code, with the similarity between the original and reconstructed code serving as an implicit supervisory signal.

Building on this ``back-translation'' strategy, we propose \textbf{BatCoder}, a self-contained framework that learns code description and code generation jointly from unlabeled code snippets. 
Given a code snippet, the model first generates a natural language document, which is then used to reconstruct the original code. 
The similarity between the original and reconstructed code provides a unified training signal that serves two complementary purposes: assessing the quality of the generated documentation and guiding the code generation process.
Recent advances in code similarity metrics, such as CSSG~\cite{xu2026cssg}, make this design practically feasible.
These similarity-based signals are incorporated as rewards within a reinforcement learning algorithm, enabling the joint optimization of both generation stages without relying on external documentation or stronger teacher models.

Our main contributions are summarized as follows:
\begin{itemize}
\vspace{-2mm}
\setlength{\itemsep}{0pt}
\setlength{\parsep}{0pt}
\setlength{\parskip}{0pt}
\item We propose, a self-supervised back-translation framework, named BatCoder, which enables bidirectional generation learning between code and documentation without relying on externally curated paired data, alleviating the scarcity of code-documentation supervision.
\item We empirically demonstrate that BatCoder exhibits favorable scaling behavior with respect to both model capacity and training data size, showing that reconstruction-based self-supervision provides increasingly effective learning signals as scale increases.
\item Through extensive experiments across multiple programming languages, we show that BatCoder consistently improves performance on both code generation and documentation production tasks, with particularly strong gains in low-resource languages.
\end{itemize}

\begin{figure*}[t]
    \centering
    \vspace{-5pt}
    \includegraphics[width=0.98\textwidth]{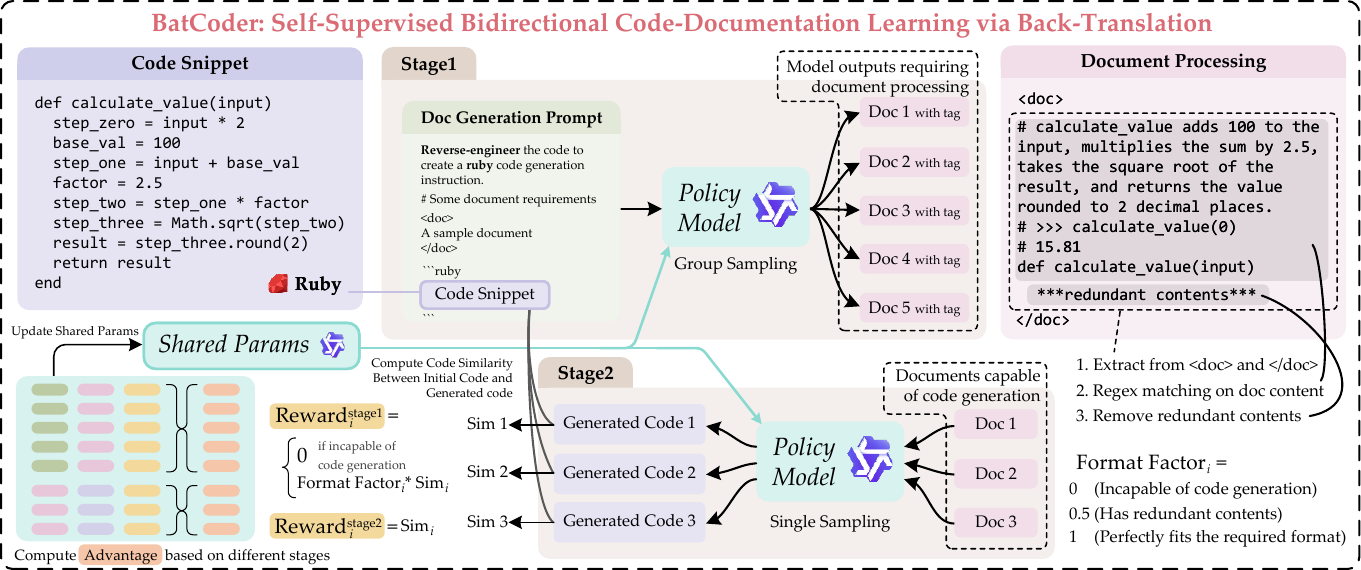}
    \caption{BatCoder training pipeline via self-supervised back-translation. Given an unlabeled code snippet $c$, the model first generates multiple documentation candidates in \emph{Stage~1} (code-to-documentation). After extracting content and filtering for structural validity, each valid candidate is used to sample a single reconstructed code in \emph{Stage~2} (documentation-to-code). Rewards consist of code similarity and document format compliance, allocated differently to the two stages. Both directions are jointly optimized via Reinforce++ algorithm.}
    \label{fig:main}
    \vspace{-5pt}
\end{figure*}

% \vspace{-2mm}
\section{Related Work}
\label{sec:related_work}

\noindent \textbf{LLMs for Code.} \quad
Modern LLMs, having absorbed massive collections of  natural language and code during training, exhibit impressive capabilities in various coding tasks, such as code generation \cite{chen2021evaluating,zhuo2024bigcodebench,jain2024livecodebench}, code summarization  \cite{lu2021codexglue, sun2025source,su2025context} and program repair  \cite{xia2023automated,jiang2023impact,jimenez2024swe}. Specialized models, such as CodeT5+ \cite{wang2023codet5+}, CodeLlama \cite{roziere2023code}, DeepSeek-Coder \cite{guo2024deepseek},
StarCoder2 \cite{Lozhkov2024StarCoder2A} and Qwen2.5-Coder \cite{hui2024qwen2}, undergo dedicated pre-training or fine-tuning on large-scale code corpora to build robust code generation and understanding capabilities.

\noindent \textbf{Data Augmentation for Code-Related Tasks.} \quad
Despite abundant raw code data, high-quality code-text pairs for model training remain scarce, motivating data augmentation techniques to construct task-specific supervision. WizardCoder~\cite{luo2023wizardcoder} iteratively evolves code-text pairs by increasing instruction complexity, incorporating constraints, and introducing adversarial elements to improve data quality. Magicoder~\cite{wei2024magicoder} synthesizes coding problems from unlabeled code snippets by composing code fragments and leveraging a strong LLM to ensure task coherence. Similarly exploiting unlabeled code, \citet{gao2024learning} generates pseudo-labels using pre-trained models and filters low-quality instances via normalized edit distance.

Closely related to our work, UniCoder~\cite{sun2024unicoder} also seeks to align source code with its semantic meaning by introducing structured intermediate representations, such as rule-constrained pseudo-code or documentation as a latent bridge between code understanding and generation. However, the quality of these intermediate representations is primarily ensured through hand-crafted rules or external stronger model judgments.

Another closely related work is SelfCodeAlign~\cite{wei2024selfcodealign}, which also explores leveraging only the base model and unlabeled code snippets. Specifically, it extracts coding concepts from seed functions, generates instruction-response pairs via in-context learning, and validates the responses through self-generated tests and sandbox execution. However, their approach mainly serves as a data augmentation strategy, as the model is trained using supervised fine-tuning on these execution-validated pairs. 

In contrast, our method employs a self-contained bidirectional transformation between code and documentation. Rather than relying on external supervision, we evaluate the quality of generated documentation by measuring how faithfully the original code can be reconstructed through a back-translation process. This back-translation similarity then serves as the primary learning signal for both directions, enabling joint optimization without reliance on rule-based heuristics or stronger external models.

\noindent \textbf{Reinforcement Learning in Code Generation.} \quad
Reinforcement learning enhances code generation by leveraging execution feedback to optimize model performance on various programming tasks. CodeRL \cite{le2022coderl} integrates pre-trained models with actor-critic structure, employing unit test rewards to bolster functional correctness. Similarly, PPOCoder \cite{shojaee2023execution} advances this paradigm via Proximal Policy Optimization (PPO), incorporating rewards from compiler feedback, syntactic AST similarity, semantic DFG matching, and KL-divergence regularization to align generations with target codes. \citet{ye2025process} enhances code generation by applying process supervision, providing denser step-wise rewards to mitigate sparse outcome signals. Complementing these, advancements in math and code reasoning\cite{liu2025acereason} synergize supervised fine-tuning with reinforcement learning, leveraging feedback from dual domains to enhance logical capabilities.

Our framework extends these approaches by deriving rewards from code reconstruction similarity within a back-translation process, eliminating the need for target code and facilitating reward propagation to the code-to-documentation stage, alongside an evaluation-train separation mechanism that alleviates memory overheads in conventional RL setups.

\section{Methods}
\label{sec:approach}
In this section, we present the formulation and training procedure of BatCoder. We first define the problem setting and learning objectives, followed by the sampling strategy used during training, the reward design based on reconstruction similarity, and the reinforcement learning algorithm for parameter optimization. BatCoder treats code description and code generation as a back-translation process, where documentation is generated from code and subsequently used for code reconstruction. Prior studies have shown that reconstruction-based similarity provides an effective signal for assessing code-documentation alignment~\citep{allamanis2024unsupervised, sharma2024patched}. Building on this insight, we use the similarity between the original code and its reconstruction as the core learning signal to jointly optimize both transformation stages. Figure~\ref{fig:main} presents an overview of the BatCoder training process.

\subsection{Problem Formulation}
Given a code snippet $ c \in \mathcal{C} $, where $ \mathcal{C} $ denotes the space of code, the BatCoder framework trains the model in two sequential stages. The transformation from code to documentation is referred to as \emph{Stage~1}, in which the model generates a descriptive document $ d = f_\theta(c) \in \mathcal{D} $, where $ \mathcal{D} $ denotes the space of natural language documentation. The inverse transformation from documentation back to code is referred to as \emph{Stage~2}, where the model reconstructs a code snippet $ c' = g_\theta(d) \in \mathcal{C} $.

The composition of \emph{Stage~1} and \emph{Stage~2}, formalized as $ g_\theta \circ f_\theta: \mathcal{C} \to \mathcal{C} $, defines a reconstruction objective that implicitly regularizes both code generation and documentation synthesis through structural faithfulness. The model parameters $ \theta $ are optimized to maximize the expected reward over the two-stage transformation:
\begin{equation}
\small
J(\theta) = \mathbb{E}_{c \sim p(c)} \left[ R(c, d, c') \right],
\end{equation}
where $ R $ evaluates syntactic and semantic similarity among $ c $, $ d $, and $ c' $, and $ p(c) $ denotes the empirical distribution of the training code corpus.

Based on the two-stage formulation, BatCoder is trained by constructing complete back-translation trajectories from code to documentation and back to reconstructed code, and jointly optimizing both stages using reinforcement learning. Each trajectory starts from an unlabeled code snippet and yields learning signals through reconstruction similarity, allowing the model to improve both documentation generation and code reconstruction without external supervision.

We next describe how training trajectories are sampled, followed by the reward formulation and the reinforcement learning algorithm used for parameter optimization.

\subsection{Sampling Strategy}
\label{sec:sampling}

During training, BatCoder adopts an asymmetric sampling strategy for the two stages. In \emph{Stage~1}, given an input code snippet, we sample $K$ documentation candidates to account for the inherent diversity of natural language descriptions. In \emph{Stage~2}, each selected documentation instance is used to generate only a single reconstructed code sample. The prompts used in \emph{Stage~1} are provided in Appendix~\ref{app:prompt}. 

This asymmetric design treats \emph{Stage~1} and \emph{Stage~2} as a single continuous rollout. For each training code snippet, we generate $K$ complete trajectories from code to documentation and back to reconstructed code. Each trajectory provides rewards for both stages, ensuring that model updates reflect the dual objectives of producing valid documentation and improving code reconstruction quality. Sampling a single reconstruction per documentation instance in \emph{Stage~2} maintains a balanced number of trajectories for both stages and additionally reduces computational and memory overhead. Before entering \emph{Stage~2}, the documentation generated in \emph{Stage~1} is filtered and rewritten according to the predefined format and validity constraints. The filtering  and rewriting criteria are described in Appendix~\ref{app:filtering}. 

Formally, the asymmetric sampling strategy produces two sets of trajectories.
In \emph{Stage~1}, for each input code snippet $c$, we sample a set of documentation trajectories
\begin{equation}
\small
\mathcal{T}_{\text{code2doc}}(c) = \{ (c, d_k) \}_{k=1}^{K},
\end{equation}

Each documentation instance $d_k$ is then subjected to the filtering and rewriting process.
For documentation instances that satisfy the validity constraints, a second-stage trajectory is constructed in \emph{Stage~2} as
\begin{equation}
\small
\mathcal{T}_{\text{doc2code}}(c) = \{ (d_m, c'_m) \}_{m=1}^{M},
\end{equation}
where $M \leq K$.

\subsection{Reward Design}
\label{sec:reward}
To enable self-supervised optimization in the absence of paired supervision, we design rewards that reflect both code-level similarity and documentation quality. Since code-level similarity can only be evaluated after reconstructing code from documentation, we first define the reward for \emph{Stage~2}, and then derive the reward for \emph{Stage~1} based on reconstruction outcomes and documentation validity.

\paragraph{Documentation-to-Code Reward.}
For \emph{Stage~2}, the primary learning signal comes from the similarity between the original code and its reconstruction generated from the documentation.
Let $\mathcal{S}(\cdot, \cdot)$ denote a code-level similarity function that measures the similarity between two code snippets.
This function is abstract and can be instantiated using different structural or semantic comparison metrics.
In our experiments, we instantiate $\mathcal{S}$ using CSSG~\cite{xu2026cssg}, a code similarity metric based on an improved program dependence graph (PDG).
CSSG produces scores in the range $[0,1]$, where higher values indicate stronger semantic and structural similarity between code snippets.

For each trajectory $(d_m, c'_m) \in \mathcal{T}_{\text{doc2code}}(c)$ generated in \emph{Stage~2}, we define a documentation-to-code similarity reward by comparing the reconstructed code with the original input code $c$:
\begin{equation}
\small
R_{\text{doc2code}}^{(m)} = R_{\text{sim,doc2code}}^{(m)} = \mathcal{S}(c, c'_m).
\end{equation}

No additional validity constraints are imposed at this stage, as the reconstruction directly reflects whether the documentation preserves the essential program semantics.

\paragraph{Code-to-Documentation Reward.}
For each trajectory $(c, d_k) \in \mathcal{T}_{\text{code2doc}}(c)$ generated in \emph{Stage~1}, we compute a similarity-based reward by comparing the original code with its reconstruction:
\begin{equation}
\small
R_{\text{sim,code2doc}}^{(k)} = R_{\text{sim,doc2code}}^{(k)} = \mathcal{S}(c, c'_k).
\end{equation}

In addition to reconstruction similarity, we incorporate a documentation validity reward $R_{\text{doc}}^{(k)}$.
Beyond reflecting the structural correctness and well-formedness of the generated documentation, this reward is designed to stabilize the documentation generation process during training by encouraging consistent adherence to the required documentation format and reducing variance in early-stage optimization.
The auxiliary validity reward is defined as follows:
\begin{equation}
\small
R_{\text{doc}}^{(k)} =
\begin{cases}
0,   & \text{incapable of code generation}, \\
0.5, & \text{contains redundant content}, \\
1,   & \text{perfectly fits the required format}.
\end{cases}
\end{equation}

The final reward for \emph{Stage~1} is defined at the trajectory level:
\begin{equation}
\small
R_{\text{code2doc}}^{(k)} = R_{\text{sim,code2doc}}^{(k)} \cdot R_{\text{doc}}^{(k)}.
\end{equation}
where $R_{\text{sim,code2doc}}^{(k)}$ is only computed when a valid reconstruction $c'_k$ is available.

In practice, the documentation generated in \emph{Stage~1} is first subjected to the filtering and rewriting process described in Section~\ref{sec:sampling}.
If a documentation instance $d_k$ fails to satisfy the predefined validity constraints, it is excluded from the subsequent documentation-to-code stage and no reconstruction is performed.
Such trajectories terminate at $(c, d_k)$ and do not contribute to the \emph{Stage~2} optimization. Since no valid reconstruction is available in this case, the code-to-documentation reward is explicitly defined as $R_{\text{code2doc}}^{(k)} = 0$.

\subsection{Reinforcement Learning Algorithm}

\begin{algorithm}[t]
\small
\caption{\textbf{BatCoder}: Self-Supervised Bidirectional Code-Documentation Learning via Back-Translation}
\label{alg:BatCoder}
\begin{algorithmic}
\STATE \textbf{Input:} Unlabeled code corpus $\mathcal{C}$, model parameters $\theta$, replay buffer $\mathcal{B}$,
iterations $T$, documentation samples $K$
\STATE Initialize $\theta$, $\mathcal{B} \leftarrow \emptyset$

\FOR{$t = 1$ to $T$}
    \STATE Sample code snippet $c \sim \mathcal{C}$
    \STATE Sample documentation candidates $\{d^{(1)}, \dots, d^{(K)}\} \sim \pi_\theta(d \mid c)$
    
    \FOR{$k = 1$ to $K$}
    \STATE Apply filtering and rewriting to $d^{(k)}$, and compute documentation validity reward $R_{\text{doc}}^{(k)}$
    
    \IF{$d^{(k)}$ fails validity constraints}
        \STATE Set $R_{\text{code2doc}}^{(k)} \leftarrow 0$
        \STATE Store $(c, d^{(k)}, R_{\text{code2doc}}^{(k)})$ in $\mathcal{B}$
        \STATE \textbf{continue}
    \ENDIF
    
    \STATE Sample reconstructed code $c'^{(k)} \sim \pi_\theta(c' \mid d^{(k)})$
    
    \STATE Compute similarity reward:
    \STATE \quad $R_{\text{sim}}^{(k)} = \mathcal{S}(c, c'^{(k)})$
    
    \STATE Set rewards:
    \STATE \quad $R_{\text{code2doc}}^{(k)} = R_{\text{sim}}^{(k)} \cdot R_{\text{doc}}^{(k)}$
    \STATE \quad $R_{\text{doc2code}}^{(k)} = R_{\text{sim}}^{(k)}$
    
    \STATE Store $(c, d^{(k)}, R_{\text{code2doc}}^{(k)})$ in $\mathcal{B}$
    \STATE Store $(d^{(k)}, c'^{(k)}, R_{\text{doc2code}}^{(k)})$ in $\mathcal{B}$
\ENDFOR
    
    \STATE Sample minibatches from $\mathcal{B}$
    \STATE Update $\theta$ using policy gradients for both directions
\ENDFOR

\STATE \textbf{Output:} Optimized model parameters $\theta$
\end{algorithmic}
\end{algorithm}

To optimize BatCoder, we adopt Reinforce++~\cite{hu2025reinforce++}, a policy gradient algorithm that supports on-policy updates for improved sample efficiency. 
Training proceeds over trajectories generated by the bidirectional code-documentation back-translation process described in Section~\ref{sec:sampling}. The reward for the \emph{Stage~2}, denoted as $R_{\text{doc2code}}$, and the reward for the \emph{Stage~1}, denoted as $R_{\text{code2doc}}$, are computed as defined in Section~\ref{sec:reward}. 
Samples generated during training are stored in a fixed-size replay buffer $\mathcal{B}$, including
code-to-documentation pairs $(c, d_k)$ with their corresponding rewards $R_{\text{code2doc}}^{(k)}$,
and documentation-to-code pairs $(d_m, c'_m)$ with rewards $R_{\text{doc2code}}^{(m)}$.

Model parameters $\theta$ are updated by sampling minibatches from $\mathcal{B}$.
To stabilize training, rewards are normalized using statistics computed over the buffer, yielding the following token-level advantage estimates:
\begin{equation}
\small
A_{\text{code2doc}}^{(k)} = \frac{R_{\text{code2doc}}^{(k)} - \mu_{\text{code2doc}}}{\sigma_{\text{code2doc}}},
\end{equation}
\begin{equation}
\small
A_{\text{doc2code}}^{(m)} = \frac{R_{\text{doc2code}}^{(m)} - \mu_{\text{doc2code}}}{\sigma_{\text{doc2code}}},
\end{equation}
where $\mu_{\text{code2doc}}, \mu_{\text{doc2code}}$ and
$\sigma_{\text{code2doc}}, \sigma_{\text{doc2code}}$ denote the mean and standard deviation of the corresponding rewards maintained in $\mathcal{B}$.

The training objective jointly optimizes both generation directions.
Specifically, the loss function is defined as follows:
\begin{align} \small
L(\theta) & =
- \mathbb{E}_{\mathcal{T} \sim \mathcal{B}}
\big[ A_{\text{code2doc}}^{(k)} \cdot \log \pi_\theta(d_k \mid c) \big] \nonumber \\
& - \mathbb{E}_{\mathcal{T} \sim \mathcal{B}}
\big[ A_{\text{doc2code}}^{(m)} \cdot \log \pi_\theta(c'_m \mid d_m) \big] \\
& + \beta \cdot \mathrm{KL}(\pi_\theta \,\|\, \pi_{\text{ref}}), \nonumber
\end{align}
where the KL regularization term constrains updates with respect to a reference policy $\pi_{\text{ref}}$.
This on-policy optimization strategy enables efficient reuse of past trajectories and promotes stable learning across both stages of the cycle. The overall procedure is summarized in Algorithm~\ref{alg:BatCoder}.

\begin{table*}[t]
\centering
\small
\setlength{\tabcolsep}{4mm}
\caption{Pass@1 (\%) results on HumanEval(+) and MBPP(+). Results for Qwen2.5-Instruct and BatCoder are obtained using the \texttt{bigcode-evaluation-harness} framework, while all other baseline results are retrieved from the EvalPlus Leaderboard. BatCoder consistently improves upon its base model and surpasses open-source baselines with comparable or even larger parameter scales.
}
\label{tab:main_experiment}
\begin{tabular}{l l c cc c}
\toprule
\multirow{2}{*}{\textbf{Scale}} &
\multirow{2}{*}{\textbf{Model}} &
\multirow{2}{*}{\textbf{Size}} &
\multicolumn{2}{c}{\textbf{Benchmark}} &
\multirow{2}{*}{\textbf{Open-Source}} \\
\cmidrule(lr){4-5}
 &  &  & \textbf{HumanEval (+)} & \textbf{MBPP (+)} &  \\
\midrule
\multirow{2}{*}{Unknown}
& GPT-4o & Unknown & 92.7 (87.2) & 87.6 (72.2) & \color{red}$\times$ \\
& O1 Mini & Unknown & \textbf{96.3} (\textbf{89.0}) & \textbf{93.1} (\textbf{78.8}) & \color{red}$\times$ \\
\midrule
\multirow{8}{*}{$>$6B}
& CodeT5+ & 16B & 31.7 (26.8) & 56.6 (47.1) & \color{blue}$\checkmark$ \\
& StarCoder2 & 15B & 46.3 (37.8) & 78.0 (65.1) & \color{blue}$\checkmark$ \\
& CodeLlama-Instruct & 34B & 51.8 (43.9) & 69.3 (56.3) & \color{blue}$\checkmark$ \\
& WizardCoder-Python & 34B & 73.2 (64.6) & 75.1 (63.2) & \color{blue}$\checkmark$ \\
& Magicoder-S-DS & 6.7B & 76.8 (71.3) & 79.4 (69.0) & \color{blue}$\checkmark$ \\
& DeepSeek-Coder-Instruct & 33B & 81.1 (\underline{75.0}) & \underline{80.4} (\textbf{70.1}) & \color{blue}$\checkmark$ \\
& Qwen2.5-Instruct & 7B & \underline{81.7} (73.2) & 78.6 (68.0) & \color{blue}$\checkmark$ \\
& \textbf{BatCoder (Ours)} & 7B & \textbf{83.5} (\textbf{76.8}) & \textbf{81.0} (\underline{69.3}) & \color{blue}$\checkmark$ \\
\midrule
\multirow{4}{*}{$\le$3B}
& StarCoder2 & 3B & 31.7 (27.4) & 57.4 (47.4) & \color{blue}$\checkmark$ \\
& DeepSeek-Coder-Instruct & 1.3B & 65.9 (60.4) & 65.3 (54.8) & \color{blue}$\checkmark$ \\
& Qwen2.5-Instruct & 3B & \underline{73.8} (\underline{68.3}) & \underline{73.0} (\underline{62.2}) & \color{blue}$\checkmark$ \\
& \textbf{BatCoder (Ours)} & 3B & \textbf{76.2} (\textbf{71.3}) & \textbf{75.9} (\textbf{66.4}) & \color{blue}$\checkmark$ \\
\bottomrule
\end{tabular}
\end{table*}

\section{Experimental Setup}
\subsection{Base Models and Training Data}

Qwen2.5-3B-Instruct and Qwen2.5-7B-Instruct~\cite{qwen2.5} were used as the base models. These models are widely adopted in recent code-related studies and allow us to evaluate the robustness of our approach across different model scales. All experiments were initialized from the corresponding pretrained checkpoints released on the Hugging Face platform.

For training, we used the code data from the Code-Text task in the CodeXGLUE benchmark suite\footnote{\url{https://huggingface.co/datasets/google/code_x_glue_ct_code_to_text}}~\cite{husain2019codesearchnet, lu2021codexglue}.
This dataset contains code samples from multiple programming languages, including Python, Ruby, and Go, making it well suited for training on a unified corpus and evaluating generalization across languages. Although the original dataset includes accompanying text fields, these texts primarily consist of function-level docstrings rather than the structured documentation targeted in our setting. Consequently, we discarded the text component and used only the code snippets during training.

\subsection{Baselines}
We compare BatCoder against a diverse set of representative baselines, including widely adopted code-oriented LLMs that are pre-trained or fine-tuned on large-scale code corpora. The open-source baselines include CodeT5+~\cite{wang2023codet5+}, CodeLlama~\cite{roziere2023code}, WizardCoder~\cite{luo2023wizardcoder}, Magicoder~\cite{wei2024magicoder}, StarCoder2~\cite{Lozhkov2024StarCoder2A}, DeepSeek-Coder-Instruct~\cite{guo2024deepseek}, and Qwen2.5-Instruct~\cite{hui2024qwen2}. In addition, we report results from two closed-source models, GPT-4o and O1 Mini, which currently represent the state-of-the-art on the EvalPlus leaderboard\footnote{\url{https://evalplus.github.io/leaderboard.html}} ~\cite{liu2023your} and serve as upper-bound references.

For each model family, we evaluate variants whose parameter scales or reported performance levels are closest to those of BatCoder, enabling a fair and meaningful comparison across comparable settings. All results are consistently drawn from the EvalPlus leaderboard.

\subsection{Benchmarks and Evaluation Protocol}

We evaluate model performance on HumanEval~\cite{chen2021evaluating} and MBPP~\cite{austin2021program}, two widely adopted benchmarks for assessing code generation quality. HumanEval and MBPP consists of Python programming problems with unit tests for execution-based evaluation. To more thoroughly assess model performance, we further report results on HumanEval+ and MBPP+, the rigorized versions from EvalPlus~\cite{liu2023your} that augment the original benchmarks with 80$\times$ and 35$\times$ more test cases.

In addition to Python-centric evaluation, we further examine model performance on programming languages with comparatively smaller training corpora, where the scarcity of paired code-documentation data is more pronounced. To this end, we include MultiPL-E~\cite{cassano2022multipl} for evaluation and focus on Ruby and Go, which are also covered in the CodeXGLUE training data. MultiPL-E extends HumanEval and MBPP by providing translated versions of these benchmarks in multiple programming languages.

We used \texttt{bigcode-evaluation-harness} framework~\cite{bigcode-evaluation-harness} for all the evaluations, which standardizes test execution and pass@1 computation across benchmarks. We employ greedy decoding (i.e., temperature = 0) with a maximum generation length of 1024 tokens.

\subsection{Training Settings}

The 3B and 7B BatCoder models are trained with Reinforce++, implemented on a modified version of the \texttt{verl} ~\cite{sheng2024hybridflow} framework (v0.5.0.dev).
Experiments are conducted on two NVIDIA A100 GPUs.

We set the number of documentation samples to $K=8$, with a maximum response length of 1500 tokens.
The training batch size is 64, the actor learning rate is $1\times10^{-6}$, and the mini-batch size is 32.
Although standard Reinforce++ incorporates KL regularization to prevent excessive policy deviation, we do not apply an explicit KL constraint in practice, as it was observed to overly restrict policy updates in our experimental setting. This can equivalently be viewed as setting the KL coefficient hyperparameter $\beta$ to 0. We additionally adopt a dynamic sampling strategy \cite{yu2025dapo}, where minibatches with zero rewards for all trajectories are discarded and excluded from parameter updates.

For ablation experiments with SFT, we use the default configuration, with a batch size of 16, a learning rate of $1\times10^{-5}$, and a maximum sequence length of 2048.

\section{Results and Discussion}
\subsection{Python Documentation-to-Code Generation}

Table~\ref{tab:main_experiment} summarizes the main results on HumanEval, HumanEval+, MBPP and MBPP+. 
At the 7B scale, BatCoder achieves consistent gains over the base Qwen2.5-Instruct model, with pass@1 improving from 81.7 to 83.5 on HumanEval and from 73.2 to 76.8 on HumanEval+. The same trend holds on the MBPP benchmarks, increasing from 78.6 to 81.0 on MBPP and from 68.0 to 69.3 on MBPP+. Notably, the 7B BatCoder model outperforms the substantially larger DeepSeek-Coder-Instruct (33B) on HumanEval, HumanEval+, and MBPP, demonstrating the effectiveness of the proposed self-supervised training signal beyond merely increasing model size.
Similar trends are observed at smaller scales. At the 3B scale, BatCoder consistently outperforms the corresponding Qwen2.5-Instruct baseline, achieving pass@1 gains of +2.4 on HumanEval, +3.0 on HumanEval+, +1.1 on MBPP, and +4.2 on MBPP+. The presence of improvements at both 3B and 7B scales indicates that BatCoder generalizes across different model capacities, rather than being tailored to a specific parameter regime.

\subsection{Multilingual Code Generation}

\begin{table}[h]
\small
\setlength{\tabcolsep}{4mm}
\centering
\caption{Pass@1 (\%) results on MultiPL-E for low-resource programming languages, including Ruby and Go. We compare the base Qwen2.5-Instruct models with BatCoder at different model scales, observing consistent and substantial improvements across both programming languages and model sizes.
}
\label{tab:multipl_results}
\begin{tabular}{l c cc}
\toprule
\textbf{Model} & \textbf{Size} & \textbf{Ruby} & \textbf{Go} \\
\midrule
Qwen2.5-Instruct & 7B & 3.1 & 34.4 \\
BatCoder (Ours)    & 7B & \textbf{13.0} & \textbf{39.0} \\
\midrule
Qwen2.5-Instruct & 3B & 0.0 & 33.8 \\
BatCoder (Ours)    & 3B & \textbf{10.6} & \textbf{37.7} \\
\bottomrule
\end{tabular}
\end{table}

We further conducted experiments on MultiPL-E for Ruby and Go, two languages for which high-quality code-documentation resources are comparatively scarce. Compared to Python, obtaining paired data for these languages is more challenging under conventional training paradigms, and the low baseline scores suggest that the base models may have limited exposure to these languages during training. This setting provides a meaningful testbed for examining whether BatCoder can not only strengthen existing code capabilities but also elicit non-trivial performance in previously underrepresented languages.
 
Table~\ref{tab:multipl_results} reports the evaluation results. BatCoder consistently outperforms the corresponding Qwen2.5-Instruct baselines across both Ruby and Go, with markedly larger gains on Ruby. In particular, at the 3B scale, the base model attains a pass@1 of 0.0 on Ruby, indicating a complete failure to solve any test instances, whereas BatCoder raises performance to 10.6. This substantial jump from zero to a non-trivial accuracy highlights the effectiveness of the proposed framework in extremely low-resource regimes. A similar but less extreme pattern is observed at the 7B scale on Ruby, where BatCoder improves pass@1 from 3.1 to 13.0, yielding an absolute gain of nearly 10 points.
On Go, where the base models already exhibit moderate performance, BatCoder still delivers consistent improvements at both model scales, increasing pass@1 from 33.8 to 37.7 for the 3B model and from 34.4 to 39.0 for the 7B model.

Taken together, these results demonstrate that BatCoder generalizes effectively across model scales and is particularly well suited to low-resource programming languages. By leveraging code-level similarity as a self-supervised signal, the framework enables meaningful performance improvements in settings where curated code-documentation pairs are scarce or unavailable, suggesting promising applicability to a broader range of low-resource code domains.

\begin{table}[t]
\setlength{\tabcolsep}{6mm}
\centering
\small
\caption{Pass@1 (\%) ablation results on Ruby from MultiPL-E. We compare SFT, partial variants of BatCoder, and the full framework, with the full model achieving the best performance.
}
\label{tab:ablation_ruby}
\begin{tabular}{l c}
\toprule
\textbf{Training Setting} & \textbf{Pass@1} \\
\midrule
Base Model       & 0.0 \\
SFT      & 6.2 \\
BatCoder (w/o \emph{Stage~1}) & 1.9 \\
BatCoder       & \textbf{10.6} \\
\bottomrule
\end{tabular}
\end{table}

\begin{figure*}[h]
\centering
\begin{subfigure}{0.32\textwidth}
    \centering
    \includegraphics[width=\textwidth]{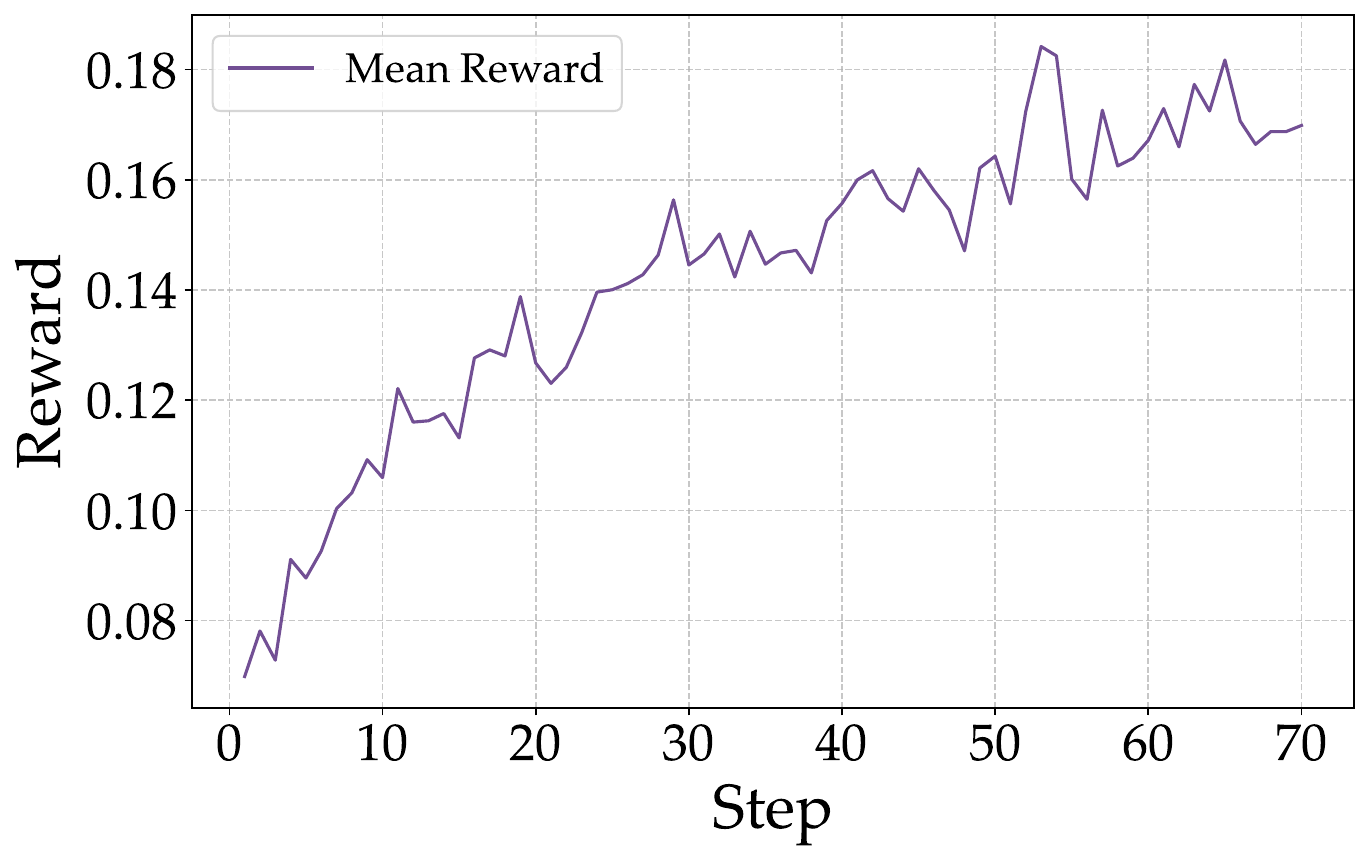}
    \caption{\emph{Stage~1} mean reward curve}
    \label{fig:stage1}
\end{subfigure}%
\hfill
\begin{subfigure}{0.32\textwidth}
    \centering
    \begin{overpic}[width=\textwidth]{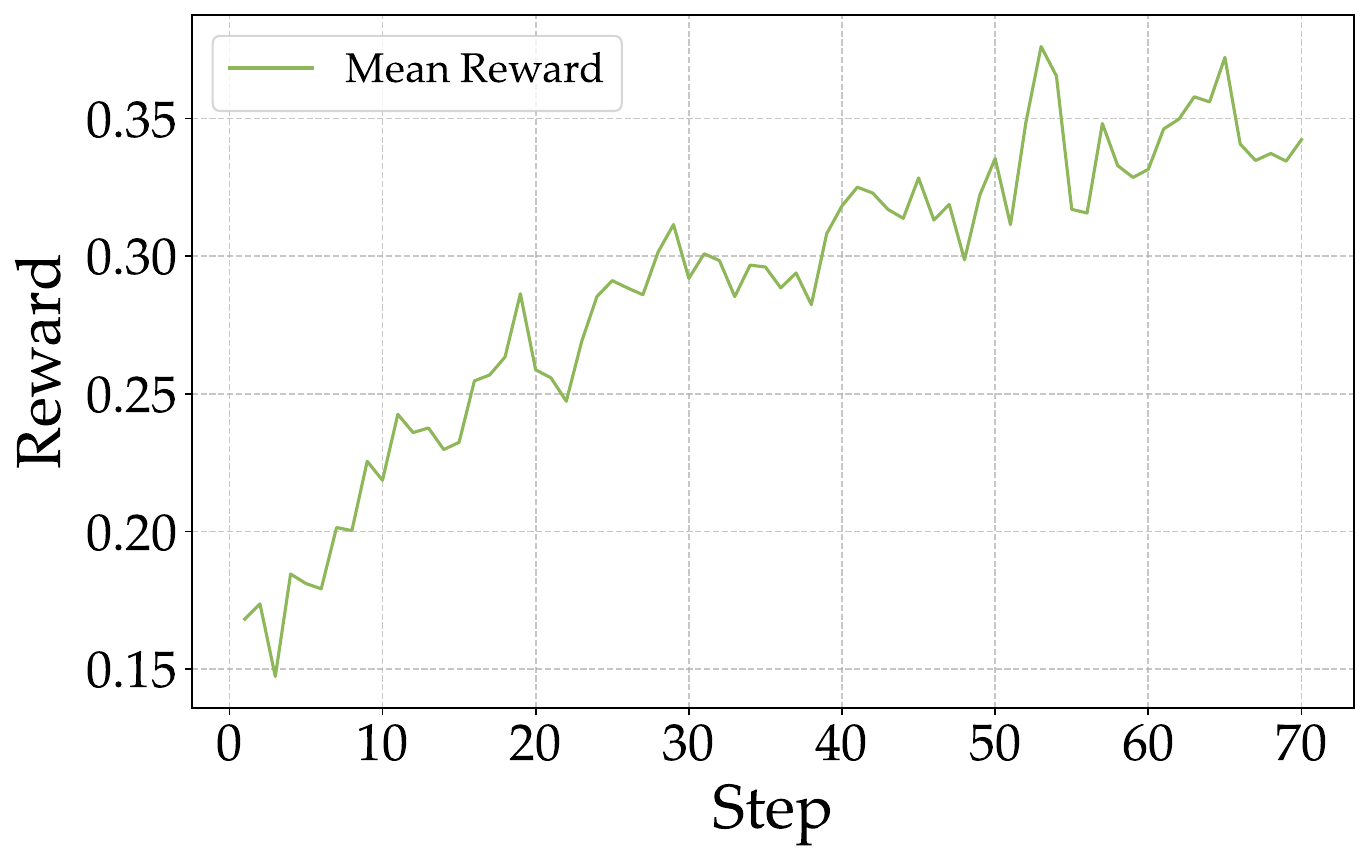}
    \end{overpic}
    \caption{\emph{Stage~2} mean reward curve}
    \label{fig:stage2}
\end{subfigure}%
\hfill
\begin{subfigure}{0.32\textwidth}
    \centering
    \begin{overpic}[width=\textwidth]{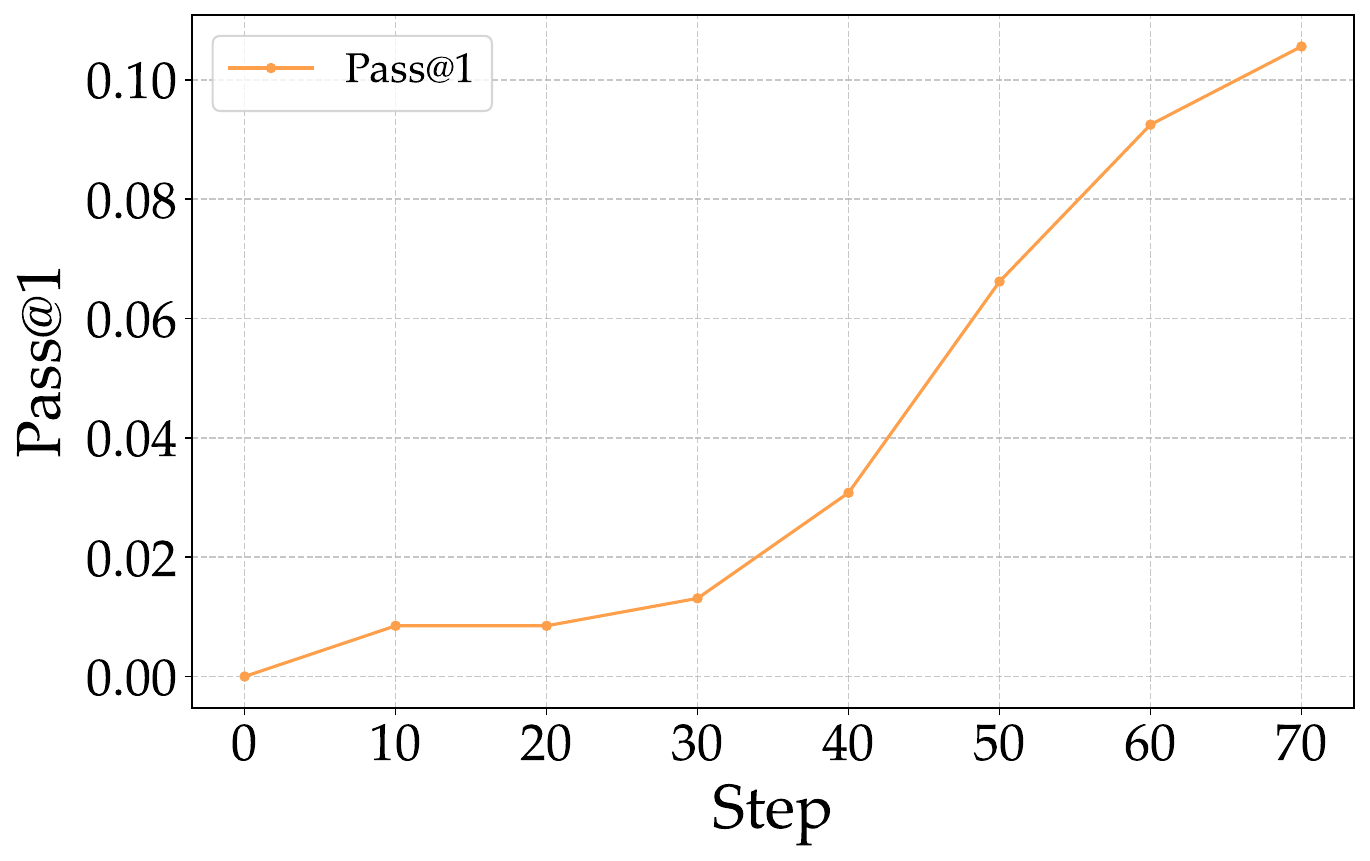}
    \end{overpic}
    \caption{pass@1 results of MultiPL-Ruby}
    \label{fig:multipl-ruby}
\end{subfigure}
\caption{Training dynamics of BatCoder. 
All curves show a consistent upward trend, indicating that the reward signals are aligned with the training objectives and correlate with improved model performance.}
\label{fig:training_dynamics}
\end{figure*}

\subsection{Ablation Study}

We conduct ablation experiments on Ruby from MultiPL-E to better understand the contribution of different components in BatCoder, with a particular focus on the role of the \emph{Stage~1} and its associated reward. The base model used in these experiments is Qwen2.5-3B-Instruct.

\paragraph{Effect of the Code-to-Documentation Stage}
We examine the impact of optimizing documentation generation by disabling parameter updates associated with \emph{Stage~1}, while preserving \emph{Stage~2} optimization based on code-level similarity. In this setting, the model still produces documentation, but it is no longer guided by an explicit reinforcement signal and does not contribute to parameter updates. As reported in Table~\ref{tab:ablation_ruby}, this variant leads to only a marginal improvement over the base model, with pass@1 increasing from 0.0 to 1.9, and remains far below the performance achieved by the complete BatCoder. This observation highlights the importance of directly optimizing documentation generation. When documentation is not encouraged to preserve sufficient information for faithful code reconstruction, the resulting self-supervised signal becomes substantially weaker, limiting the effectiveness of subsequent code generation optimization. To illustrate the differences between documentation generated before and after BatCoder training, we provide a case study in Appendix~\ref{app:case_study}.

\paragraph{Comparison with Supervised Fine-Tuning.}
We further compare BatCoder against a SFT baseline to assess whether the observed gains can be attributed solely to exposing the model to additional synthetic data. In this setting, documentation is generated by the base model, with the same filtering and rewriting procedures applied as in BatCoder, and resampling performed until valid documents are obtained. The model is then fine-tuned using these documents as inputs and the original code as targets, following a conventional SFT paradigm. While SFT improves performance over the base model, it remains notably inferior to the complete BatCoder. This suggests that training on model-generated pseudo pairs is insufficient. Instead, explicitly evaluating and reinforcing documentation quality through code-level similarity provides a more effective learning signal.

\subsection{Training Dynamics and Reward Analysis}

To analyze the training dynamics induced by the proposed reinforcement learning formulation, we track the evolution of the mean rewards for \emph{Stage~1} and \emph{Stage~2} throughout training, together with the downstream Pass@1 performance on MultiPL-Ruby. All curves are obtained from the training trajectories of BatCoder-3B on the Ruby subset. Figure~\ref{fig:training_dynamics} summarizes these dynamics across training checkpoints.

As shown in Figures~\ref{fig:stage1} and \ref{fig:stage2}, the mean rewards for both stages increase steadily as training progresses, exhibiting smooth upward trends. Figure~\ref{fig:multipl-ruby} reports the corresponding Pass@1 scores on MultiPL-Ruby, which also improve consistently over training steps and follow a similar progression pattern to the reward curves.

Overall, the consistent alignment between reward trajectories and downstream evaluation performance suggests that the proposed reward design provides a stable and meaningful optimization signal. The bidirectional reinforcement learning objective effectively guides both documentation generation and code reconstruction, resulting in tangible improvements in code generation quality. Although our analysis is conducted at a fixed training scale, the observed training dynamics indicate that the framework has the potential to further benefit from larger training data.

\section{Conclusion and Future Work}

We present \textbf{BatCoder}, a self-supervised framework that jointly learns code generation and documentation synthesis through a back-translation paradigm. By enforcing consistency between code and documentation under bidirectional transformations, BatCoder enables effective optimization directly from unlabeled code, without requiring paired supervision or external teacher models.

Experimental results show that BatCoder consistently improves code generation performance across multiple benchmarks, outperforming  supervised and synthetic-data-based baselines at comparable model scales. Notably, the proposed framework is particularly effective in low-resource programming languages, such as Ruby and Go, where curated code-documentation pairs are scarce. These findings indicate that back-translation similarity provides a robust learning signal in data-constrained settings, facilitating meaningful improvements from unlabeled code alone. 

Several directions are promising for future work. One direction is to incorporate more diverse reward signals and evaluate them under alternative reinforcement learning algorithms to assess the robustness of similarity-based rewards. Another avenue is to further investigate scaling behavior along multiple dimensions, including larger training corpora, increased model capacity, and alternative architectural choices, to better understand the regimes in which BatCoder is most effective. In addition, the proposed framework may be extended to related tasks such as code completion or code translation. The limitations are discussed in Appendix~\ref{app:limitation}.

\section*{Accessibility}
We have made efforts to ensure that this submission is as accessible as possible to a broad audience, including readers with disabilities or sensory and neurological differences.

\section*{Software and Data}
We have shown all the prompts in the Appendix, and we will release the core code of our approach on Github upon acceptance.

\section*{Impact Statement}

This work focuses on self-supervised learning for code generation and documentation synthesis via a back-translation learning process. We demonstrate consistent improvements on standard benchmarks, low-resource programming languages, and across varying data and model scales. We do not think our work will negatively impact ethical aspects or future societal consequences.

\nocite{langley00}

\bibliography{main}
\bibliographystyle{icml2026}

%%%%%%%%%%%%%%%%%%%%%%%%%%%%%%%%%%%%%%%%%%%%%%%%%%%%%%%%%%%%%%%%%%%%%%%%%%%%%%%
%%%%%%%%%%%%%%%%%%%%%%%%%%%%%%%%%%%%%%%%%%%%%%%%%%%%%%%%%%%%%%%%%%%%%%%%%%%%%%%
% APPENDIX
%%%%%%%%%%%%%%%%%%%%%%%%%%%%%%%%%%%%%%%%%%%%%%%%%%%%%%%%%%%%%%%%%%%%%%%%%%%%%%%
%%%%%%%%%%%%%%%%%%%%%%%%%%%%%%%%%%%%%%%%%%%%%%%%%%%%%%%%%%%%%%%%%%%%%%%%%%%%%%%
\newpage
\appendix
\onecolumn
\section{Prompt Templates}
\label{app:prompt}

This appendix presents the prompt templates used for documentation sampling in \emph{Stage~1}. The prompts are designed to elicit structured and well-formed documentation that can be reliably used for subsequent reconstruction in \emph{Stage~2}. We adopt a one-shot prompting strategy, providing a single format-aligned example to encourage the model to consistently follow the desired documentation structure.

\subsection{Python Documentation Generation Prompt}

\begin{lstlisting}
Please analyze the code provided at the end and reverse-engineer it to create a python code generation instruction.
The output must be wrapped in `<doc>` and `</doc>` tags and include:

1. Any necessary library imports.
2. The Python function definition line with type annotations.
3. Indented by 4 spaces, a docstring starting with `"""` that includes a description of what the code does.
4. Indented by 4 spaces, one or several illustrative input/output examples (using `>>>` syntax) inside the docstring.
5. Indented by 4 spaces, the closing `"""` of the docstring.

Here is an example of the expected structure (pay attention to the format, not the content):

<doc>
import json
def test_func(arg1: str, arg2: int) -> bool:
    """ test_func implements the functionality of ...
    >>> test_func('example', 1)
    True
    """
</doc>

Now, please generate the python code problem statement within `<doc>` and `</doc>`:
```{origin_lan}
{code}
```
\end{lstlisting}

\subsection{Ruby Documentation Generation Prompt}

\begin{lstlisting}
Please analyze the code provided at the end and reverse-engineer it to create a ruby code generation instruction.
The output must be wrapped in `<doc>` and `</doc>` tags and include:

1. Any necessary `require` statements.
2. A descriptive problem statement comment explaining what the code does.
3. One or several illustrative input/output comment line (using `>>>` syntax).
4. The Ruby function definition line matching the logic (needs to complete generating the leading spaces of the next line).

Here is an example of the expected structure (pay attention to the format, not the content):

<doc>
require 'json'
# test_func implements the functionality of ...
# >>> test_func(arg1, arg2)
# expected_result
def test_func(arg1, arg2)
  </doc>

Now, please generate the ruby code problem statement within `<doc>` and `</doc>`:
```{origin_lan}
{code}
```
\end{lstlisting}

\subsection{Go Documentation Generation Prompt}

\begin{lstlisting}
Please analyze the code provided at the end and reverse-engineer it to create a go code generation instruction.
The output must be wrapped in `<doc>` and `</doc>` tags and include:

1. The package declaration and relevant imports.
2. A descriptive problem statement comment explaining what the code does.
3. One or several illustrative input/output comment line (using `>>>` syntax).
4. The Go function definition line matching the logic (needs to complete generating the leading tab of the next line).

Here is an example of the expected structure (pay attention to the format, not the content):

<doc>
package main
import "fmt"
// test_func implements the functionality of ...
// >>> test_func(arg1, arg2)
// expected_result
func test_func(arg1 Type, arg2 Type) Type {
	</doc>

Now, please generate the go code problem statement within `<doc>` and `</doc>`:
```{origin_lan}
{code}
```
\end{lstlisting}

\section{Filtering and Rewriting}
\label{app:filtering}

This appendix describes the filtering and rewriting procedure applied to the documentation generated in \emph{Stage~1} before it is passed to \emph{Stage~2}. The purpose of this process is to retain only structurally valid and semantically informative documentation, ensuring that reconstruction-based rewards are computed on reliable inputs under the back-translation setting.

\subsection{Documentation Structure Constraints}

To enforce a consistent documentation format, we apply a set of structural constraints using regular-expression-based matching. These constraints are aligned with the prompt specification and the one-shot example provided in Appendix~\ref{app:prompt}, ensuring consistency between the expected documentation format and the filtering criteria.

A documentation instance is considered valid if it contains:
(i) one or more natural language description lines summarizing the functionality,
(ii) at least one illustrative input-output example, and
(iii) a function or procedure definition corresponding to the described behavior.
These constraints are designed to be language-agnostic and capture common documentation patterns across programming languages.

\subsection{Filtering and Rewriting Procedure}

The filtering procedure consists of three steps.
First, only the content enclosed within designated documentation tags (e.g., \texttt{<doc>} and \texttt{</doc>}) is extracted from the raw model output. Generations without valid tags are discarded.
Second, the extracted content is matched against the structural constraints described above. If a valid match is found, the documentation is truncated at the end of the matched span, removing any trailing or extraneous content beyond the required structure. This rewriting step is applied only when additional text appears after the end of the structurally valid documentation. In such cases, the instance is marked as containing redundant content, and only the prefix that satisfies the structural constraints is retained.

In addition to producing a cleaned documentation instance, the filtering process records whether the generated documentation terminates exactly at the matched boundary, which serves as an auxiliary indicator of structural completeness. Only documentation that passes this filtering and rewriting procedure is used as input to \emph{Stage~2}.

\section{Case Study: Documentation Quality Comparison}
\label{app:case_study}

To qualitatively analyze the differences  before and after BatCoder training, we present a case study comparing the generated documentation for the same code snippet after training with the two approaches. The goal of this analysis is to examine whether BatCoder leads to higher-quality and more informative documentation, even though no explicit reward is imposed directly on documentation content.

The following Ruby helper function is used as the input code for documentation generation:

\begin{lstlisting}
def sign_in(user = create(:user))
  visit new_user_session_path
  within('#new_user') do
    fill_in 'user_email', with: user.email
    fill_in 'user_password', with: user.password
  end
  click_button 'Log in'
end
\end{lstlisting}

This function implements an end-to-end user login procedure in a Rails system test, including navigation to the login page, form filling, and submission.

\paragraph{Base Model Generated Documentation.}
The documentation generated by the base model is shown below.

\begin{lstlisting}
require 'capybara'
require 'capybara/dsl'
require 'support/database_cleaner'

# sign_in generates the login process for a user.
# >>> sign_in(user: create(:user))
# This will log in a user with the specified email and password.
def sign_in(user = create(:user))
  <end>
\end{lstlisting}

\paragraph{BatCoder Generated Documentation.}
In contrast, the documentation produced by BatCoder is as follows:

\begin{lstlisting}   
require 'rails_helper'
# The `sign_in` method is used to simulate a user signing in to the application.
# It visits the login path, fills in the email and password for the given user,
# and submits the form by clicking the log in button.
# >>> sign_in(create(:user))
# visits the new_user_session_path, fills in email and password, and clicks the Log in button
def sign_in(user = create(:user))
  <end>
\end{lstlisting}

\paragraph{Analysis.}
Under the filtering and rewriting mechanisms introduced in Appendix~\ref{app:filtering}, both outputs satisfy the required documentation format and are therefore considered valid. However, their quality differs substantially.

The base model produces only a coarse and generic description of the function, largely restating that it performs a login operation without capturing the procedural structure or key execution steps. In contrast, BatCoder generates a more detailed and functionally grounded description that explicitly reflects the control flow of the code, including page navigation, form interaction, and submission behavior.

Although BatCoder does not impose an explicit reward on documentation content itself, higher-quality documentation improves the fidelity of the subsequent code reconstruction stage. More precise descriptions enable more accurate back-translation, which in turn yields higher code-level similarity rewards. This indirect supervision effectively guides the model toward generating more informative and semantically aligned documentation, demonstrating the advantage of the proposed bidirectional reinforcement learning framework over the unaligned base model.

\section{Limitations}
\label{app:limitation}
While BatCoder demonstrates strong empirical performance across multiple benchmarks and programming languages, we acknowledge several limitations that point toward promising avenues for future work.

First, we used the same set of hyperparameters for both 3B and 7B model training without extensive tuning. While this simplifies the experimental setup, our results already demonstrate the effectiveness of BatCoder, suggesting even stronger performance is achievable with tailored hyperparameter optimization.

Second, the reward signals in \emph{Stage~1} and \emph{Stage~2} rely solely on code similarity and documentation formatting. We did not explore alternative similarity metrics, their weighting schemes, or additional reward sources. Nevertheless, the consistent gains across benchmarks indicate substantial headroom for improvement through richer reward design.
\end{document}